\documentclass[10pt]{article}

\usepackage{amsmath}
\usepackage{amssymb}

\usepackage{cite}

\usepackage{hyperref}
\usepackage{graphicx}

\usepackage{lineno}

\usepackage{microtype}

\usepackage[linesnumbered,noend, noline, ruled]{algorithm2e}
\usepackage{subfigure}

\usepackage{setspace} 

\usepackage[labelfont=bf,labelsep=period,justification=raggedright]{caption}

\makeatletter
\renewcommand{\@biblabel}[1]{\quad#1.}
\makeatother

\date{}
\begin{document}

\begin{flushleft}
{\Large
\textbf{A Context-aware Delayed Agglomeration Framework for Electron Microscopy Segmentation}
}
\\
Toufiq Parag$^{1, \ast}$, 
Anirban Chakrobarty$^{2}$, 
Stephen Plaza$^{1}$,
Lou Scheffer$^{1}$
\\
{\bf 1} Janelia Farm Research Campus, HHMI, Ashburn, VA, USA.
\\
{\bf 2} University of California, Riverside, VA, USA
\\
$\ast$ E-mail: paragt@janelia.hhmi.org
\end{flushleft}

\section*{Abstract}
Electron Microscopy (EM) image (or volume) segmentation has become significantly important in recent years as an instrument for connectomics. This paper proposes a novel agglomerative framework for EM segmentation. In particular, given an over-segmented image or volume, we propose a novel framework for accurately clustering regions of the same neuron. Unlike existing agglomerative methods, the proposed context-aware algorithm divides superpixels (over-segmented regions) of different biological entities into different subsets and agglomerates them separately. In addition, this paper describes a \lq delayed\rq~ scheme for agglomerative clustering that postpones some of the merge decisions,  pertaining to newly formed bodies, in order to generate a more confident boundary prediction. We report significant improvements attained by the proposed approach in segmentation accuracy over existing standard methods on 2D and 3D datasets.


\section{Introduction}
Extracting the network structure among neurons in animal brain has gained substantial attention lately in the field of neuroscience. Rapid advances in imaging technology, in particular Electron Microscopy (EM) techniques, have enabled us to trace neural bodies in unprecedented level of details. However, recording in such high resolution (at nanometer scale) generates massive amount of data that is too large to annotate manually. Automated region labeling or segmentation is considered to be the most viable strategy for generating a dense reconstruction of neural anatomy. Some recent efforts of such reconstruction yielded impressive results utilizing machine learning/computer vision tools such as image segmentation, and offered valuable biological insights to the neuroscience community~\cite{helmstaedter13nature}\cite{takemura13}. 


Image segmentation for natural scenes has a long history in computer vision literature~\cite{arbelaez11}\cite{krahenbuhl11}\cite{ladicky09}. In recent years, there have also been many fruitful attempts to automatically identify meaningful regions in EM images using segmentation techniques~\cite{andres08}\cite{andres12}\cite{chklovskii10}\cite{funke12}\cite{jain11}\cite{vazquez11}. Most of these studies initially apply a pixelwise\footnote{\scriptsize Unless it is obvious from the context, we denote locations on both 2D and 3D EM data as pixels in this manuscript.} classifier to determine whether or not any particular pixel belongs to the cell boundary. The quantified confidence values of the pixelwise classifier are utilized to produce an initial (over-) segmentation through methods such as Watershed~\cite{meyer93}. 

Different approaches resort to different methods to generate the final segmentation by merging or clustering the over-segmented bodies to corresponding neuronal cells. Andres et.al.~\cite{andres12} addresses this problem by searching the optimal subset of superpixel borders that form closed surfaces. Several studies work with the watershed merge tree in order to identify the regions to be combined for the final segmentation~\cite{jurrus12icpr}\cite{uzunbas14}. Some approaches applied  agglomerative or hierarchical clustering ~\cite{chklovskii10}\cite{jain11}\cite{nunez13} for this purpose. For anisotropic datasets, where the depth resolution ($z$-dimension) is coarser than the planar resolution ($x,y$ dimensions), segmented bodies on one section overlap with multiple regions in the adjacent sections. Therefore, a complete 3D reconstruction need to establish the correct correspondence, through an alignment or co-segmentation technique~\cite{funke12}\cite{vazquez11}, among segmented regions across multiple planes . In this study, we restrict ourselves only to segmentation on images for anisotropic data -- the subsequent alignment is out of the scope of this paper. For both isotropic and anisotropic reconstruction, the outputs of segmentation algorithms need to be  corrected afterwards, either manually\cite{takemura13} or by combining with a manually traced skeletonized representation~\cite{helmstaedter13nature}.

Biologically, the interior of a neuron cell comprises several distinct sub-structures (or sub-categories) such as cytoplasm, mitochondria, vesicles etc. An ideal binary pixel classifier --  which assigns a pixel to one of the two categories: cell boundary and cell interior -- should label all locations within these sub-categories to cell-interior class. Several past studies~\cite{ciresan12}\cite{jain10cvpr}\cite{jurrus12mia}\cite{liu14} recommend increasingly complex  pixelwise detector models to attain a binary prediction output. In contrast, some recent works~\cite{nunez13}\cite{vazquez11} represent the sub-categories (cytoplasm, mitochondria etc.) of cell body by multiple classes and apply relatively simpler classifiers (in terms of model size, learning time and convenience) for this multiclass classification problem. The results of~\cite{nunez13}\cite{vazquez11} suggest that using prior domain knowledge to divide a problem into multiple components can achieve high segmentation quality with simpler classifier models requiring less computation.

\begin{figure}[h]
\begin{center}
\subfigure[\scriptsize  grayscale image]{\includegraphics[width=0.325\columnwidth, height=0.3\columnwidth]{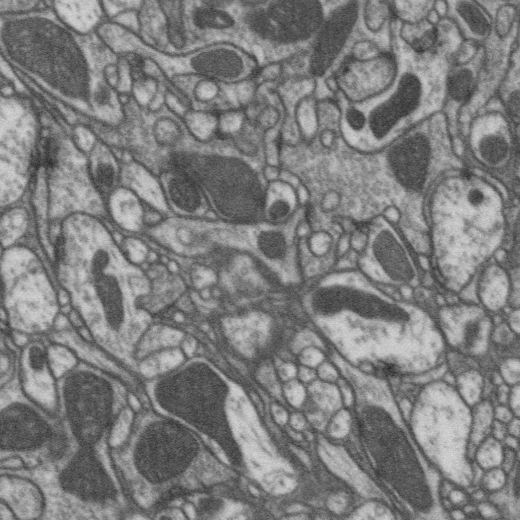}} \qquad
\subfigure[\scriptsize mitochondria probability]{\includegraphics[width=0.325\columnwidth, height=0.3\columnwidth]{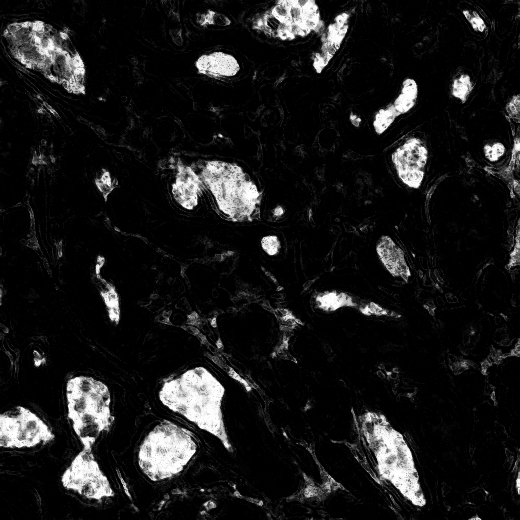}}\\
\subfigure[\scriptsize output of context oblivious method GALA.]{\includegraphics[width=0.325\columnwidth, height=0.3\columnwidth]{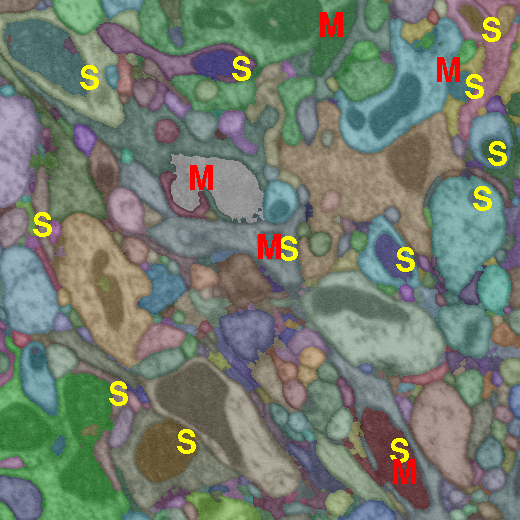}}\qquad
\subfigure[\scriptsize output proposed method]{\includegraphics[width=0.325\columnwidth, height=0.3\columnwidth]{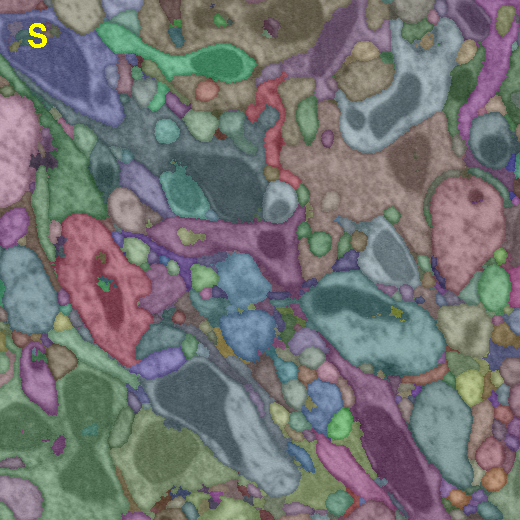}}
\vspace{-0.4cm}\caption[\bf Improved segmentation accuracy by context-aware approach on FIBSEM data.]{ \scriptsize Improved segmentation accuracy by context-aware approach on FIBSEM data. (a) one plane of input volume, (b) mitochondria detection on that plane, (c) the output of GALA~\cite{nunez13} (context oblivious), and (d)the output of proposed context aware method. The segmented region labels are overlaid on the image using random artificial colors. S and M on images indicate locations of false split and merge respectively}\label{F:INTRO_MITO}
\end{center}
\end{figure}

However, we believe the methods of~\cite{nunez13}\cite{vazquez11} do not exploit the full benefit of multiclass predictions on EM data. Regardless of the quality of multiclass pixel classifier output, the algorithms in~\cite{nunez13}\cite{vazquez11} do not distinguish between regions of one sub-structure (e.g., cytoplasm) from those of another (e.g., mitochondria) at the superpixel level. That is, the classification is divided into multiple sub-classes in pixel-level, but the subsequent fusion or superpixel clustering step does not utilize this additional information to compute the final segmentation. This often leads to sub-optimal performances by these methods. For example, Figure~\ref{F:INTRO_MITO} shows an EM image (plane in a 3D volume) and the corresponding pixel predictions for the mitochondria sub-class. By not using this sub-class prediction explicitly in the clustering step, the final output of~\cite{nunez13} failed to merge many regions into the correct cell (marked by \lq S\rq) and connected some of them to wrong cells (marked by \lq M\rq).  

%


This paper introduces a context-aware scheme for combining over-segmented regions by utilizing the prior knowledge of sub-classes. We adopt an agglomerative or hierarchical clustering framework ~\cite{chklovskii10}\cite{jain11} due to its advantages such as low space, time complexity and flexibility to tune for over/under segmentation.  We develop a two-pass agglomeration policy where the (estimated) cytoplasm regions are grouped together in the first phase and then the remaining mitochondria bodies are absorbed into the cell cytoplasm. In these two stages, the superpixels are agglomerated based on different merge criteria  that are defined by different contexts, which is why we call it context-aware agglomeration. Our proposed context aware approach significantly reduces the false split and merge errors (example shown in Figure~\ref{F:INTRO_MITO}) provided fairly accurate sub-structure detection. In addition, this strategy substantially reduces the training data requirement, as well as the predictor model complexity, which in turn offers significant increase in learning speed.  The findings of this study further inspired us to design an interactive training algorithm~\cite{miccai14interactive} for region boundary predictor that does not require exhaustively labeled groundtruth. Generating such an exhaustive annotation is considered to be a bottleneck for neural reconstruction~\cite{helmstaedter13}.


We also propose a modified version of the hierarchical clustering algorithm to cluster the superpixels in both phases of the context-aware framework. The proposed clustering method emphasizes on minimizing under-segmentation errors since these errors are conventionally costlier to correct than the over-segmentation errors~\cite{chklovskii10}. In order to minimize the number of false merges, we \lq delay\rq~ the merge decisions on a certain type of boundaries to be resolved at a later time. Compared to the traditional agglomerative scheme of~\cite{jain11}, the proposed modification reduces the number of false merges significantly. We also attempt to analyze why our agglomeration approach performs better than the Global multicut scheme~\cite{andres12} on the dataset used for our experiments.

The paper is organized as follows. We define the problem in Section~\ref{S:METHODS} and briefly describe the existing clustering segmentation algorithms in Section~\ref{S:PRIOR}. Then we explain the proposed delayed agglomeration scheme in Section~\ref{S:DELAYED}. This delayed strategy is employed in both the stages of our context-aware algorithm discussed in Section~\ref{S:TWOSTAGE}. Section~\ref{S:RESULT} reports our experimental setup, both quantitative and qualitative results and their analyses. We conclude and discuss our findings further in Section~\ref{S:DISCUSSION}.


\section{Methods}\label{S:METHODS}
A formal definition of the problem we are addressing assumes an initial over-segmentation,   comprising $N$ superpixels $ \{ S_1, S_2, \dots, S_N\} \subseteq {\cal S}$, of an EM image or volume with $M$ neurites (neuronal regions) where $ N \gg M$. Let $L(S)$ be the neurite region that $S$ actually belongs to. Our goal is to correctly assign these $N$ superpixels such that each $S_i, i = 1, 2, \dots, N$ is assigned to its corresponding $L(S_i)$. 

We denote a boundary between two superpixels (i.e., oversegmented regions) by a pair of regions $e \triangleq \{ S_i, S_j\}$ and the set of all such boundaries by $E$. In a graph representation, each of the regions $S_i$ is considered to be a node and the boundary or face between two regions is regarded as an edge -- a notation we will be using throughout the paper. Also, let the boundary label map $B: {\cal S} \times {\cal S} \to \{0, 1\}$ assign a 1 to a boundary that actually separates one neurite region from another and a 0 to the boundary incorrectly generated due to over-segmentation.  The problem of correctly merging $S_i$ to its corresponding $L(S_i)$ is similar to a clustering problem where the number of clusters cannot be computed a priori. Following\cite{chklovskii10}\cite{jain11}\cite{nunez13}, we adopt an agglomerative approach for superpixel clustering.  

In our context-aware scheme, the set of superpixels is divided into two subsets: 1) the  ${\cal S}_c$ of potential cytoplasm superpixels, and 2) the set ${\cal S}_m$ of potential mitochondria superpixels. The set of cytoplasm superpixels is clustered first with the proposed delayed agglomeration algorithm. Agglomeration of the mitochondria superpixels is also performed by the proposed delayed method, but with a different merge criterion.  In order to assist the reader to comprehend the novelty of the proposed approach, we introduce the prior studies on agglomerative clustering for EM segmentation~\cite{chklovskii10}\cite{jain11}\cite{nunez13} in Section~\ref{S:PRIOR}.  Afterwards, Sections~\ref{S:DELAYED} and~\ref{S:TWOSTAGE} discuss the delayed agglomeration and the context- aware framework respectively.

\subsection{Prior Works on Agglomerative Clustering for EM Segmentation}\label{S:PRIOR}

Several existing EM segmentation approaches~\cite{chklovskii10}\cite{jain11}\cite{nunez13} tackled the problem of superpixel clustering by agglomerative hierarchical clustering, as described in Algorithm~\ref{A:AGGLO_PREV}. These methods assume a superpixel boundary estimator $h :{\cal S} \times {\cal S} \to \mathbb{R}$ that assigns real valued confidences to all edges in $E$. This boundary estimator may represent the real-valued prediction of  a classifier distinguishing true boundaries from the false ones~\cite{jain11}, or compute the mean value of boundary pixel probabilities~\cite{chklovskii10} or return the overlap percentage between borders of two adjacent superpixels. The value of $h(\{S_i, S_j\}) \in [0, 1]$ indicates how confident the estimator is about the existence of a true boundary between $S_i$ and $S_j$: a large $h(\{S_i, S_j\})$ implies the estimator is very confident that the boundary $\{S_i, S_j\}$ is correct while a small value implies the boundary was probably generated as an artifact of over-segmentation and therefore is false. Given such a function $h$,  the hierarchical clustering algorithm iteratively merges cell boundaries in the increasing order of confidence values $h(e)$ (Line~\ref{LINE:START} in Algorithm~\ref{A:AGGLO_PREV}) until a stopping criterion is satisfied, e.g., $h(e) > \delta$ where $\delta$ is a pre-defined threshold. After each merge, it also updates the neighborhood structure of the merged superpixel, i.e.,  the neighbors of the absorbed region become the neighbors of the (newly) merged cell.

\begin{algorithm}
{\small
	\KwIn{$S_1, S_2, \dots, S_N$ and confidence function $h$.}
	\KwOut{$R_1, R_2, \dots, R_{N'}$}
	\lForAll{$i$}{$R_i = S_i$}
	\Repeat{$h(e) \le \delta$}{
		$e^* \equiv \{R_i, R_j\} = min_{e \in E}~ h(e)$  \; \label{LINE:START}
		Merge $R_j$ to $R_i$ and update $E$ \;
		\ForAll{ $R_b \in \text{Nbr}(R_j) $}{
			Recompute $h(\{ R_i, R_b\})$\;  \label{LINE:UPDATE}
		}
	}
	\caption{Existing Agglomerative Segmentation}
	\label{A:AGGLO_PREV}}
\end{algorithm}

Each time a superpixel border is dissolved in standard agglomerative clustering, it modifies the characteristic representations of the pixels within the superpixels and on the boundary. This demands the confidences of the estimator function $h(e)$ on these boundaries be recomputed (Line~\ref{LINE:UPDATE} in Algorithm~\ref{A:AGGLO_PREV}). The edges, for which $h(e)$ decreases due to a merge, receive higher priority to be dissolved than it had before. The proposed delayed agglomeration strategy modifies this step and postpones merging these edges for a later time.

\subsection{Proposed Delayed Agglomerative Clustering} \label{S:DELAYED}


Our adaption of segmentation also starts with the boundary with lowest estimator confidence and repeatedly dissolve edges with in ascending order of $h(e)$. Recall that, $h(e)$ may measure the prediction of a superpixel boundary classifier, or the mean probability values on boundary pixels, or the fraction of overlap between the borders of two superpixels. After two regions have been joined due to a merge, the boundaries of the combined region is updated and the estimator function $h$ is applied to recompute the new confidences. The edges for which $h(e)$ decreases are set aside to be considered at a later stage. They are reexamined after all the borders, initially generated by over-segmentation process, have been checked.  

This method is described in Algorithm~\ref{A:AGGLO_DELAYED}.  After region $R_j$ is absorbed into $R_i$,  we do not \emph{immediately} consider all the new boundaries $\{R_i, R_b\}$ between the recently merged $R_i$ \footnote{$R_i$ is now the union of previous $R_i$ and $R_j$} and its updated neighbors $R_b$. We maintain a set of edges $W$ and insert the new edge $\{R_i, R_b\}$ only if its confidence increases from that of $\{R_j, R_b\}$ after $R_j$ is absorbed into $R_i$ (Line~\ref{LINE:UPDATE_NEW} in Algorithm~\ref{A:AGGLO_DELAYED}). The faces, for which $h(\{R_i, R_b\})$ decreases from previous value, are kept aside until there are no members left in $W$ and the modified confidence on $\{R_i, R_b\}$ is less than the agglomeration threshold (Line~\ref{LINE:PUT_OFF} in Algorithm~\ref{A:AGGLO_DELAYED}). Once all  $e \in W$ have been considered for merge and $W$ is empty, these boundaries repopulate the list $W$ (Line~\ref{LINE:REENTER} in Algorithm~\ref{A:AGGLO_DELAYED}) and renew the agglomeration process which continues until there exists no $e$ such that $h(e) \le \delta$.

Effectively, the proposed strategy \lq delays\rq~ the merging of new edges $\{R_i, R_b\}$ resulting from a merge: either due to an increase in $h(\{R_i, R_b\})$ or  deliberately if $h(\{R_i, R_b\})$ decreases. This design postpones the merge decisions on the newly formed bodies for a later time to avoid propagating wrong decisions, made on smaller superpixels,  to the larger ones. Our analyses support that deferring decisions on these edges significantly reduces false merges during agglomeration.


\begin{algorithm}[h]
{\small
	\SetKwData{Within}{within range}
	\SetKwData{Empty}{empty}
	\SetKwData{Flag}{Flag}
	\SetKwData{Region}{region}
	\SetKwData{Edge}{edge}
	\SetKwData{Delay}{DELAY}
	\KwIn{$S_1, S_2, \dots, S_N$ and confidence function $h$.}
	\KwOut{$R_1, R_2, \dots, R_{N'}$}
	\lForAll{$i$, \Edge $e$}{$R_i = S_i$, ~Flag$(e)=$ ACTIVE}
	\Repeat{$h(e) \le \delta$}{
		$W = \{ e \in E ~|~ \text{Flag}(e) = \text{ACTIVE}\}$ \;		
		\If{$W$ \Empty}{
			\ForAll{$e: \Flag(e) =$ \Delay}{
				\lIf{$h(e)$ \Within} {$\Flag(e)$ = ACTIVE }\label{LINE:REENTER}
			}			
		}
		$e^* \equiv \{R_i, R_j\} = min_{e \in W}~ h(e)$ \;\label{LINE:MIN}
		Merge $R_j$ to $R_i$, i.e., {$R_i = \{R_i \cup R_j \}$}, and update $W$ \; 
		\ForAll{ $R_b \in \text{Nbr}(R_i) $}{ 
			\eIf{$h(\{ R_i, R_b\}) > h(\{R_j, R_b\})$}{$\text{Flag}(\{ R_i, R_b\})$ = 					ACTIVE\; \label{LINE:UPDATE_NEW}} 
			{$\text{Flag}(\{ R_i, R_b\})$ = DELAY\; \label{LINE:PUT_OFF}}		
		}
	}
	\caption{Delayed Agglomerative Segmentation}
	\label{A:AGGLO_DELAYED}}
\end{algorithm}

\subsubsection{Time Complexity}\label{S:DELAYED_TIME}
Asymptotically, the running time of the delayed algorithm remains the same as the traditional agglomerative clustering in the worst case. In a priority queue implementation, instead of adding the adjacent boundaries to the queue, the delayed algorithm stores them in a separate list. Later, building a queue from this list would require $O(n_1)$ time where the  length $n_1$ of new list must be smaller than that of the previous one (which contains all edges): $n > n_1$. 

In fact, our implementation is tuned to reduce the running time of delayed agglomeration. Notice that, a subset of adjacent boundaries are not pushed back or updated into the queue (Line 13 of Algorithm~\ref{A:AGGLO_DELAYED}). We may as well apply a simple trick to avoid updates at each merge altogether: instead of increasing key of the edges with increased $h$ value (Line 11 of Algorithm~\ref{A:AGGLO_DELAYED}), we can postpone the check and increase the key until it becomes a candidate for merge (Line 7 of Algorithm~\ref{A:AGGLO_DELAYED}) or in Line 6 when it being considered to be inserted into $W$. Thus, we can reduce the computation by $O(d nlogn)$  where $d$ is the degree of $S_2$ and $n$ is the queue size.

\subsection{Proposed Context-aware Segmentation}\label{S:TWOSTAGE}

The proposed context-aware agglomeration has two different phases. We separate the set ${\cal S}_m$ of potential mitochondria superpixels from the set  ${\cal S}_c$ of potential cytoplasm superpixels assuming the existence of an effective mitochondria superpixel detector (e.g.,~\cite{lucchi12}). The regions in ${\cal S}_c$ are agglomerated first by the proposed delayed policy. Motivated by~\cite{andres08}\cite{jain11}\cite{nunez13}, a Random Forest (RF)~\cite{breiman01} classifier $h_c$ is trained to act as the boundary predictor function for clustering the set ${\cal S}_c$ of cytoplasm superpixels. During $h_c$ training, mitochondria-cytoplasm borders are treated the same way as cell membrane.


In the second step, the mitochondria-cytoplasm edges are merged in the same delayed scheme as explained in Section~\ref{S:DELAYED}, but with a different estimator function $h_m$. In order to absorb mitochondria into corresponding cells, we apply the delayed-agglomeration algorithm with a small alteration. The \emph{set of candidate edges $W$ only contains the edges between mitochondria and cytoplasm}, that is, $W = \{ \{ S_c, S_m\} ~|~ \text{type}(S_c) = \text{Cyto}, \text{type}(S_m) = \text{Mito}, \text{Flag}(\{S_c, S_m\}) = \text{ACTIVE}\}$;   \emph{ mitochondria-mitochondria edges are not considered for agglomeration}. Biologically, each mitochondrion should reside within a cell body. Therefore, boundary confidence for mitochondria merging should reflect how much a mitochondrion is contained within a cytoplasm. In order to quantify this, we define the overlap ratio $\rho(\{S_m, S_c\})$ to be the fraction of the total boundary of $S_m$ which separates $S_m$ from $S_c$ : $\rho(\{S_m, S_c\}) = {\text{length}(\{S_m, S_c\}) \over \sum_i \text{length}(\{S_m, S_i\})}$. For any edge $\{S_m, S_c\}$ with a mitochondria superpixel $S_m$ and a cytoplasm superpixel $S_c$, the confidence is defined as $h_m(\{S_m, S_c\}) = 1 - \rho(\{S_m, S_c\})$. 

\begin{figure}
\begin{center}
\subfigure[$\rho(S_m, S_c) \ge 0.8$]{\includegraphics[width=0.28\columnwidth, height=0.28\columnwidth]{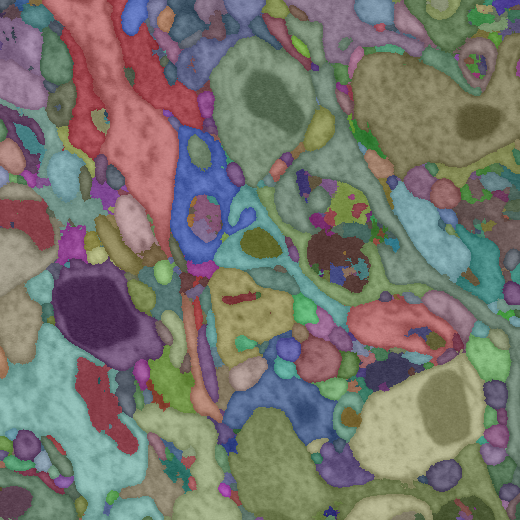}}\quad
\subfigure[$\rho(S_m, S_c) \ge 0.5$]{\includegraphics[width=0.28\columnwidth, height=0.28\columnwidth]{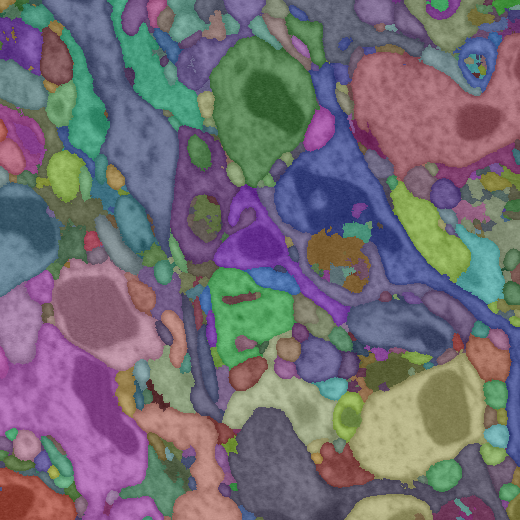}}\quad
\subfigure[$\rho(S_m, S_c) \ge 0.2$]{\includegraphics[width=0.28\columnwidth, height=0.28\columnwidth]{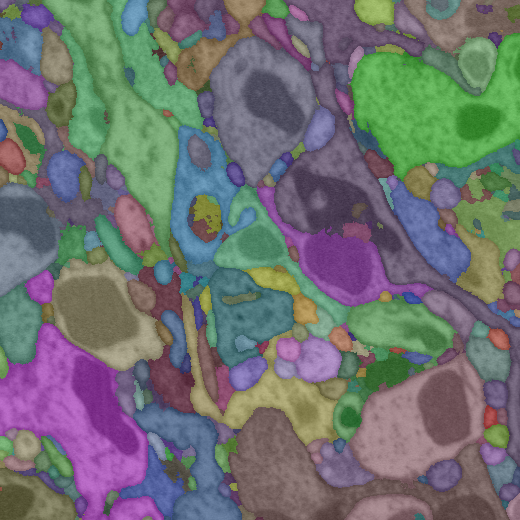}}
\vspace{-0.4cm}\caption[\bf Merging mitochondria into cytoplasm.] {\scriptsize Merging mitochondria into cytoplasm. The figure shows mitochondria superpixels absorbed into cytoplasm superpixels up to different values of overlap ratio $\rho(S_m, S_c)$}
\label{F:MITO_MERGE}
\end{center}
\end{figure}

In effect, the mitochondria superpixels are combined with the cytoplasm superpixels in the descending order of the overlap ratio between these two types of regions. That is, a mitochondria superpixel is merged into the adjacent cytoplasm region with the largest overlap between their boundaries. The combined cytoplasm-mitochondria superpixel created by such merge then identifies the next mitochondria superpixel with the largest overlap to absorb in the next step. We show snapshots of this process, at different values of $\rho(S_m, S_c)$ in Figure~\ref{F:MITO_MERGE}. It is worth noting that, for 3D segmentation, the overlap is computed across many different planes on which the two cells are neighbors to each other.
  
\section{Results}\label{S:RESULT}
We have applied the proposed method to EM images of two different modalities: isotropic Focused Ion Beam Scanning Electron Microscope (FIBSEM) data and anisotropic serial section Transmission Electron Microscopy (ssTEM) data. For both types of input data, the image (volume for the isotropic data) is first over-segmented for the agglomeration to be applied on. In the following sections, we explain our over-segmentation process and the error measures used to evaluate segmentation performance before reporting the results on FIBSEM and ssTEM data in Sections~\ref{S:RESULT_FIBSEM} and~\ref{S:RESULT_TEM} respectively.

\subsection{Over-segmentation and training:} \label{S:OVERSEG}
We learn a classifier to assign each individual pixel into multiple categories such as cell boundary, cytoplasm, mitochondria and  mitochondria boundary etc. using the interactive tool Ilastik~\cite{ilastik11}. Our pixelwise detector is  a Random Forest (RF) classifier~\cite{breiman01} trained on a few sparse samples from the dataset. The locations with lowest pixelwise  cell boundary prediction are utilized as markers for the Watershed algorithm~\cite{meyer93} to produce an over-segmentation of the image/volume. Unless otherwise specified, the same pixel prediction and watershed regions are provided as input to all (competing) methods.

The set ${\cal S}_m$ of probable mitochondria superpixels is populated  with all regions having mean mitochondria probability (estimated by our pixelwise RF classifier trained by Ilastik) above a certain threshold. The rest of the superpixels constitute the set ${\cal S}_c$ of possible cytoplasm regions. The training set for superpixel boundary classifier $h_c$ consists of all boundaries among members of ${\cal S}_c$ as well as the mitochondria-cytoplasm borders. Similar to~\cite{andres08}\cite{nunez13}, each superpixel edge is represented by the statistical properties of the multiclass probabilities estimated by Ilastik. The statistical properties include mean, standard deviation, 4 quartiles of the predictions generated for the data locations on the boundary, two regions it separates as well as the differences of these region statistics. All of these features can be updated in constant time after a merge -- a property which improves the efficiency of the segmentation algorithm substantially. The code and example dataset are publicly available at \url{https://github.com/janelia-flyem/NeuroProof.git.}

\subsection{Segmentation error measures: } We report segmentation error of both types, namely under- and over-segmentation, separately because one of these errors (under-segmentation) is costlier than the other. Split versions of variance of information (VI)~\cite{meila03} and Rand Error (RE)~\cite{jain11} were selected to evaluate segmentation errors. Given a groundtruth (GT), $GT =\{ g_1, \dots, g_M \}$, and a segmentation (SG), $SG = \{ r_1, \dots, r_P \} $, we compute the over-segmentation (OE) and under-segmentation (UE) errors by splitting the terms in VI and RE. For split-VI, the over and under-segmentation are quantified as follows.

\begin{eqnarray}
VI_{OE}= H(\text{GT} ~|~ \text{SG}) &=& - \sum_{i, j} {|g_i \cap r_j| \over Z} ~\log {{| g_i \cap r_j |} \over |g_i|}.\\
VI_{UE}= H(\text{SG} ~|~ \text{GT}) &=& - \sum_{i, j} {|r_i \cap g_j| \over Z } ~\log {{| r_i \cap g_j |} \over |r_i|}.
\end{eqnarray}
 
\noindent In these equations, $|\cdot|$ denotes the size, $\cap$ denotes the intersection between two regions and $Z$ is a normalizing constant.  From information theoretic perspective, these two terms are conditional entropies defined over a set GT given SG, and vice versa.

We also quantify segmentation error by average percentage ($\times 10^{-5}$) of pairs of voxels falsely merged and split by any method. Formally, the over-segmentation (OE) and under-segmentation (UE) is computed based on the following formula.

\begin{eqnarray}
 RE_{OE} &=& \text{\% pixel pairs within same cluster in GT but different cluster in SG.} \\
 RE_{UE} &=& \text{\% pixel pairs within same cluster in SG but different cluster in GT.}
\end{eqnarray} 

\subsection{Segmentation Performances-FIBSEM data:} \label{S:RESULT_FIBSEM}
\textbf{Dataset:} The first set of experiments was conducted on isotropic datasets from fruit fly visual system imaged at 10 nm isotropic resolution using FIBSEM technology. This data is segmented as a volume (i.e., 3D segmentation) and both the voxelwise multi-class predictor and the supervoxel boundary classifier are learned on one $250^3$ volume and applied on two $520^3$ test volumes.

\noindent \textbf{Competing methods:} We have compared the following algorithms in this study: 1) LASH: Standard agglomeration with an RF supervoxel classifier learned based on the iterative procedure of~\cite{jain11}. 2) LASH-D: LASH classifier with delayed agglomeration (proposed extension). 3)GALA~\cite{nunez13}: an agglomerative method with repetitive learning phases like LASH, except it accumulates the training sets of multiple phases. 4) CADA-F: Proposed two stage delayed agglomeration with standard RF learned using  training set accumulation similar to GALA. 5) CADA-L: Proposed delayed agglomeration with a depth-limited RF (depth =20) learned \emph{without} training set accumulation.  6) Global Multicut: the optimization framework for finding a closed-surface segmentation proposed in~\cite{andres12}. For~\cite{andres12}, the boundary confidences were generated by the CADA-L predictor.  

\begin{figure}
\begin{center}
\subfigure[Test vol1: split-VI]{\includegraphics[width=0.35\columnwidth, height=0.32\columnwidth]{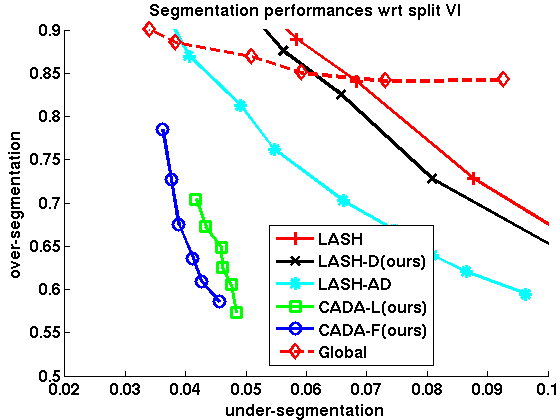}}\qquad
\subfigure[Test vol1: split-RI]{\includegraphics[width=0.32\columnwidth, height=0.32\columnwidth]{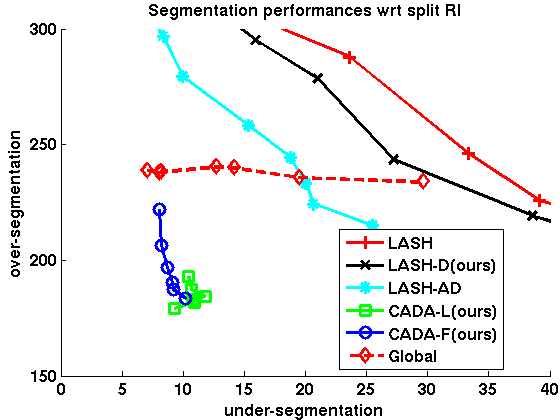}}\\
\subfigure[Test vol2: split-VI]{\includegraphics[width=0.35\columnwidth, height=0.32\columnwidth]{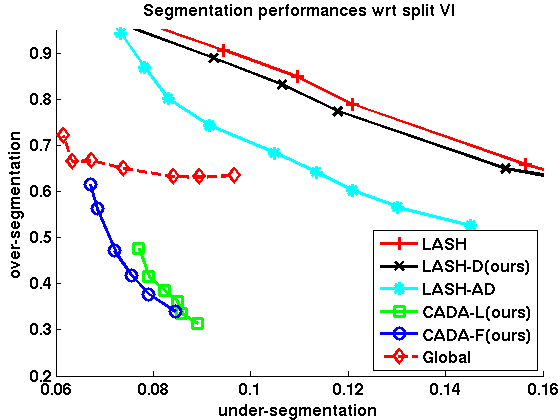}}\qquad
\subfigure[Test vol2: split-RI]{\includegraphics[width=0.32\columnwidth, height=0.32\columnwidth]{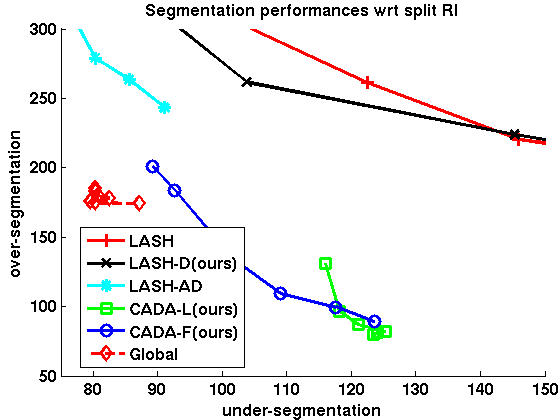}}
\end{center}
\caption[\bf Segmentation error in terms of split-VI and split-RE on two FIBSEM volumes] {\scriptsize Segmentation error in terms of split-VI and split-RE on two FIBSEM volumes. Top: Test volume 1 and bottom: Test volume 2. Left column shows split-VI error: $VI_{UE}$ in x-axis, $VI_{OE}$ in y-axis;  right column shows split-RE: $RE_{UE}$ in x-axis, $RE_{OE}$ in y-axis. Each curve is the average of results in 5 trials. Each point represents either a stopping point for clustering or bias parameter for~\cite{andres12}.}
\label{F:RESULT_Q}
\end{figure}

\noindent \textbf{Performance evaluation:} In order to compare different supervoxel clustering schemes, we trained (on one $250^3$ volume) and segmented two $520^3$ volumes 5 times and averaged their scores. We plot the average (over 5 trials) $VI_{UE}$ and $VI_{OE}$  respectively on x and y-axis respectively in plots on the left column of Figure~\ref{F:RESULT_Q} and for test Volumes 1 and 2. Similarly, we show the average $RE_{UE}$ and $RE_{OE}$ errors on x and y-axis respectively on right columns of Figure~\ref{F:RESULT_Q}. In these figures, an ideal algorithm should achieve a zero value for both over and under-segmentation. For all algorithms except the Global multicut method, each point in a plot refers to the boundary confidence threshold $\delta_c \in [0.1, 0.2]$ which was used as stopping criterion for cytoplasm merging. For~\cite{andres12}, we instead changed the value of the bias parameter in weight calculation within the range $[0.2, 0.9]$. 

As the plots show, both the delayed agglomeration and two-phase segmentation process attained significant improvement over past methods : compare the performance of LASH (red +) with LASH-D (black x) and that of GALA (cyan *) with CADA variants (green square and blue circle). Compared to the rest of the techniques, the two variants of proposed methods, namely CADA-L and CADA-F, appear to achieve the most favorable segmentations by reducing the over-segmentation steeply without increasing the false merge numbers much. During segmentation, the delayed version decreases the time needed for segmentation approximately 5 times among the agglomerative approaches. 

It is also worth mentioning that, in a two stage segmentation scheme, the performance of a depth limited RF (i.e., CADA-L, green square), learned without accumulating training set over multiple passes, is very similar to that of the standard RF (CADA-F, blue circle) trained over cumulative learning passes. Training full-depth RF (CADA-F) with multiple passes needed several hours whereas training a depth limited single iteration (CADA-L) required $\le 5$ minutes.


\begin{figure}
\begin{center}
\includegraphics[width=0.95\columnwidth, height=0.32\columnwidth]{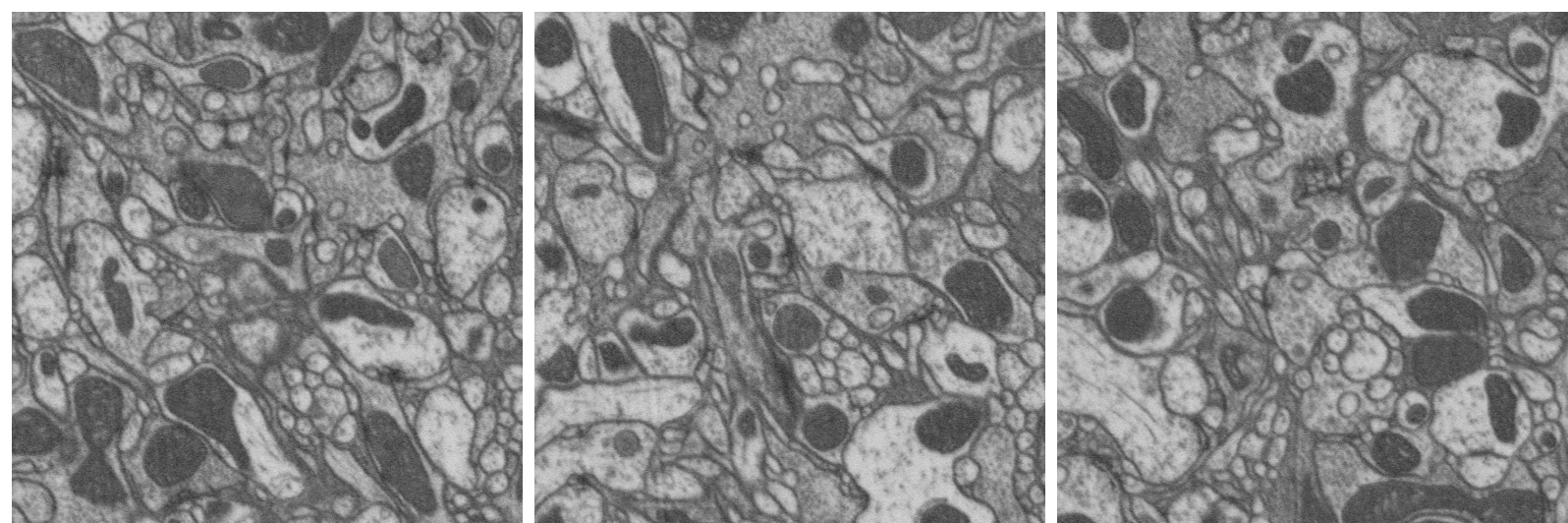}
\end{center}
\caption[\bf Sample planes from FIBSEM test volume 1.] {\scriptsize Sample planes from the test volume 1.}
\label{F:INPUT_IMG}
\end{figure}

\begin{figure}
\vspace{-1.5cm}
\begin{center}
\includegraphics[width=\columnwidth, height=1.2\columnwidth]{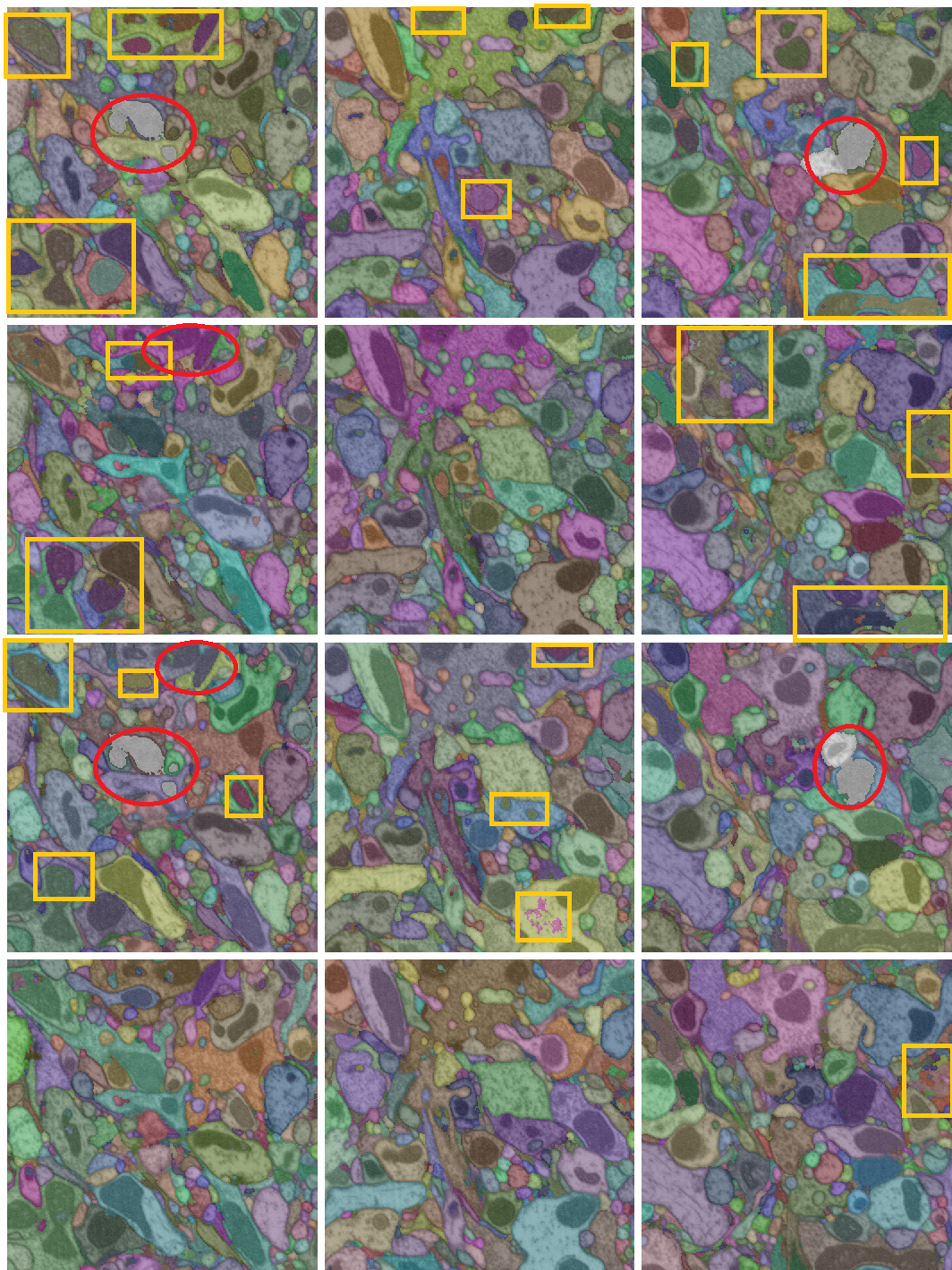}
\end{center}
\vspace{-0.5cm}
\caption[\bf Qualitative results on FIBSEM test volume 1.] {\scriptsize Qualitative results on test volume 1. Three columns show segmentation outputs overlaid with random colors on three planes of the FIBSEM volume. The rows, from top to bottom, show the output of LASH-D, Global multicut~\cite{andres12}, GALA~\cite{nunez13} and the proposed method CADA-L. Some significant over-segmentation errors and under-segmentation errors are marked in yellow rectangles and red ellipses respectively.}
\label{F:RESULT_IMG}
\end{figure}

In Figure~\ref{F:RESULT_IMG}, we show example outputs from the methods LASH-D, GALA~\cite{nunez13}, Global multicut~\cite{andres12} and CADA-L on three planes from the test volume 1 as depicted in Figure~\ref{F:INPUT_IMG}~(the algorithms were applied on 3D volume, we are showing three slices of the volume for demonstration). Three columns correspond to three planes, and each row presents the outputs of the aforementioned methods. The segmentation labeling is overlaid with artificial (randomly selected) colors. We have selected the parameter that results in the lowest false merges (under-segmentation) with a false split (over-segmentation) error below $0.7$ for all except the proposed CADA-L for which we selected the lowest over-segmentation (error value approx 0.56). The results are largely compatible with the quantitative ones, with all three methods, especially Global method, leaving many false boundaries intact. The false-splits are not limited to cytoplasm mitochondria  borders, both Global and GALA over-segmented some cytoplasm regions as well. By separating these two sub-classes within cell bodies, the proposed method CADA-L was able to eliminate the false merges between them.

\noindent \textbf{Runtime comparison:} In Table~\ref{T:RUNTIME}, we report the running times of the context oblivious standard agglomerative clustering used in LASH~\cite{jain11}; the context-oblivious delayed agglomerated clustering used in LASH-D, GALA~\cite{nunez13}; the context-aware delayed agglomeration used in CADA-L, CADA-F; and the Global Multicut~\cite{andres12} method. Both the standard and delayed agglomeration were executed up to the same threshold $\delta =0.2$. The context-aware method executes two phases of agglomeration, which is why CADA required more time than LASH-D. The Global multicut algorithm utilizes the solution of an optimization problem (requires optimization packages like CPLEX or Gurobi) in order to find the edges to merge for producing the final segmentation.
\begin{table}
\caption{Runtime for different algorithms.}  
\label{T:RUNTIME}
\begin{center}
\begin{tabular}{ |c|c| } 
 \hline
 Method & Run time (min) \\ 
 \hline
 LASH: Standard Agglomeration, context-oblivious &  $5.35 \pm 0.2$ \\ 
 LASH-D: Delayed Agglomeration, context-oblivious &  $2.72 \pm 0.06$ \\ 
 CADA: Delayed Agglomeration, context-aware & $4.69 \pm 0.02$ \\ 
 Global multicut & $7.13 \pm 1.1$ \\ 
 \hline
\end{tabular}
\end{center}
\end{table}

During the training of superpixel boundary classifier, all the algorithms except CADA-L and Global Multicut performs standard agglomeration multiple times (we repeated 5 times) in order to obtain extensive training sets. Both CADA-L and Global method exploited the same classifier learned from the initial set of boundaries existed in the over-segmented data (without training set augmentation).  


In the following subsections, we attempt to analyze why (or how) the proposed strategies improve the segmentation performance  over the existing approaches. 


\subsubsection{Context-aware vs Context-oblivious agglomeration} 
It is perhaps intuitive that traditional context-oblivious agglomeration will result in higher degree of over-segmentation than the context-aware method.  The mitochondria-cytoplasm borders indeed have strong feature similarity with cell membranes and consequently superpixel boundary predictors cannot distinguish between these two types of borders perfectly. Recall that, for segmentation,  we need to dissolve the mitochondria-cytoplasm border but retain the cell boundaries. In order to substantiate our claim, we trained a superpixel boundary classifier in context-oblivious fashion (0: false cell membrane, 1:true cell membrane) and computed its confidences on these two types of boundaries. Figure~\ref{F:INTRO_MITO} shows the histogram of confidence levels for actual cell boundaries and mitochondria-cytoplasm borders in red and blue respectively. If we wish to minimize false merges among neurons, we have to stop agglomeration at a lower value ( $\delta \le 0.3$). The overlap between the two distributions in the range $0.1~\sim0.5$ suggests that many of the mitochondria borders will not be merged and will lead to over-segmentation. 

\begin{figure}
\begin{center}
\subfigure{\includegraphics[width=0.45\columnwidth, height=0.3\columnwidth]{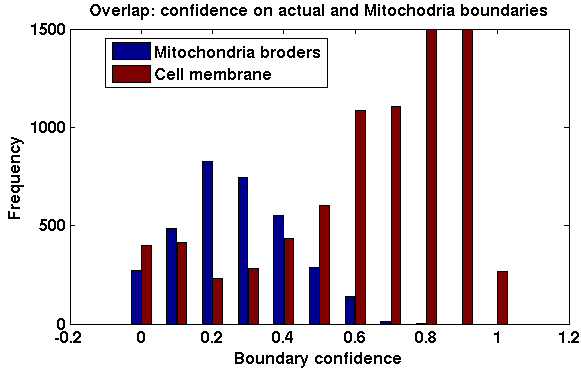}}
\end{center}
\vspace{-0.4cm}\caption[\bf Distribution of predicted confidences  on cytoplasm-mitochondria borders and  cell boundaries.]{\scriptsize Distribution of predicted boundary confidences on cytoplasm-mitochondria borders (blue) and correct cell boundaries (red). The plot is clipped at $y=1500$ for better visualization. Notice the overlap between these two distributions within confidence range $[0, 0.6].$}
\label{F:MITO}
\end{figure}

In addition, due to appearance dissimilarity, the distribution of same features computed on cytoplasm and mitochondria will be substantially different from each other. Combining these two types of feature value distribution will impede the identification of false boundaries between cytoplasm superpixels such as the one in the lower left corner of the output of GALA in Figure~\ref{F:INTRO_MITO}. 
 
In practice, mitochondria from two different cells could also lead to false merges. Often the mitochondria regions from two cells are closely located to the cell membrane, or other mitochondria regions from neighboring cells,  blurring the boundary. Figure~\ref{F:INTRO_MITO} and~\ref{F:RESULT_IMG} show several such locations where the existing techniques failed to avoid false merge. 


\subsubsection{Global multicut vs Proposed:} 
\begin{figure}
\begin{center}
\subfigure[Test vol1: split-VI]{\includegraphics[width=0.35\columnwidth, height=0.32\columnwidth]{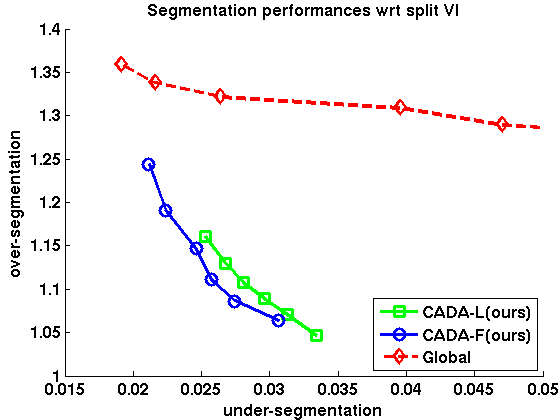}}\qquad
\subfigure[Test vol2: split-VI]{\includegraphics[width=0.35\columnwidth, height=0.32\columnwidth]{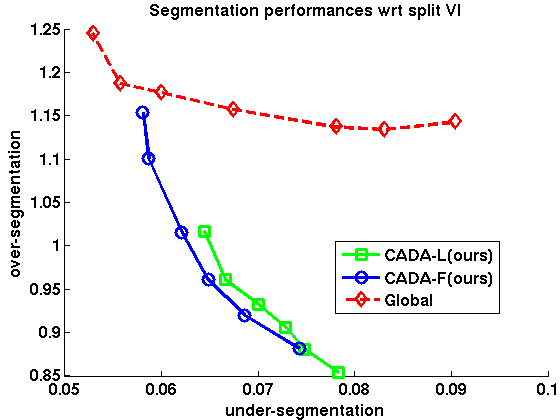}}
\vspace{-0.4cm}\caption[\bf Segmentation errors for cytoplasm only.]{ \scriptsize Split-VI of cytoplasm segmentation of two FIBSEM volumes. Left column: test volume 1, right column: test volume 2. Each curve is the average of results in 5 trials. Each point represents either a stopping point for clustering or bias parameter.}
\label{F:RESULT_Q_CYTO}
\end{center}
\end{figure}

The split-VI plot in Figures~\ref{F:RESULT_Q_CYTO} show that both variants of the proposed CADA algorithm generates significantly low under and over-segmentation errors than those of Global~\cite{andres12} method in clustering \emph{cytoplasm regions only (mitochondria not merged)}. In order to analyze why this happens, we save the initial confidences (predictor confidence at the beginning of agglomeration) of $h_c(e)$  on all $e$ that
\begin{itemize}
\item were incorrectly split (over-segmented) by the Global method, 
\item were correctly merged by proposed algorithm.
\end{itemize}

\begin{figure}
\begin{center}
\subfigure[]{\includegraphics[width=0.35\columnwidth, height=0.32\columnwidth]{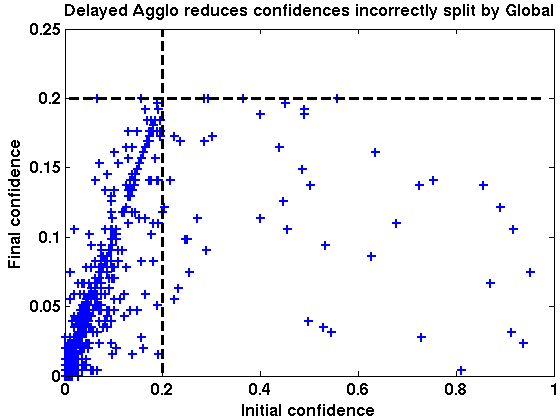}}\qquad
\subfigure[]{\includegraphics[width=0.35\columnwidth, height=0.32\columnwidth]{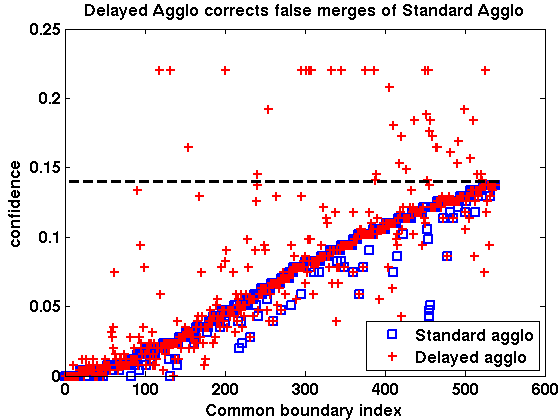}}
\vspace{-0.4cm}\caption[\bf Analyses of results.]{\scriptsize Left: False splits (over-segmentation) of Global method corrected by proposed CADA-L. Each point corresponds to a false boundary that Global method failed to dissolve. The x-axis labels indicate the predictor confidence at the beginning of the proposed agglomeration and y-axis plots the predictor confidence at the point it was merged accurately by the agglomeration. Right: False merges (under-segmentation) of Standard agglomeration corrected by delayed method -- x-axis: boundary indices, y-axis: predictor confidence. The confidences computed for the same correct edge in traditional agglomeration and in the proposed delayed version is plotted in blue square and red '+'. The confidences on many true boundaries were increased by the delayed approach.}
\label{F:RESULT_ANALYZE}
\end{center}
\end{figure}

These boundary predictions were plotted on x-axis of Figure~\ref{F:RESULT_ANALYZE}. The y-axis of Figure~\ref{F:RESULT_ANALYZE} corresponds to confidences $h_c(e)$ at the time $e$ was correctly merged by the proposed method. The threshold on boundary confidences to stop agglomeration was $\delta_c=0.2$.

Notice that, the agglomerative process correctly reduced the confidences of many false boundaries that received a high score by the predictor at the beginning (high x value but low y value). This refinement is possible through the evolution of the superpixels in the agglomerative process -- an advantage the Global method of~\cite{andres12} cannot benefit from. The Global method~\cite{andres12}, in comparison, generated many more false positive boundaries as depicted by the rectangular enclosed region of Figure~\ref{F:RESULT_ANALYZE} (left). If several boundaries within a chain of supervoxel faces receive very high predictor confidences, by construction, the  Global method tends to retain the another boundary $e$ within the same chain with a low $h_c(e)$. Such tendency may be the reason behind the high concentration of false splits with low $h_c$ within the rectangular region in Figure~\ref{F:RESULT_ANALYZE} (left).

\subsubsection{Delayed vs standard agglomeration:} In order to illustrate the improved accuracy attained by the delayed agglomeration over the standard one, we collected all \emph{faces that were  incorrectly dissolved by standard agglomeration algorithm} (LASH) and examine their confidences under a delayed scheme (LASH-D) operating at $\delta_c
 = 0.14$  The confidences (clipped to 0.25) of these 534 edges generated by standard and delayed agglomeration are plotted in Figure~\ref{F:RESULT_ANALYZE} (right) in blue square and red $+$ respectively.  The proposed delayed agglomeration accurately increased the confidences $h_c$ of many of these faces, among which, 41 exceeded the threshold of $0.14$ (green line) and avoided a false merge. In addition to these common supervoxel edges, the standard and delayed algorithms independently generated 163 and 4 more incorrect merges respectively.

\subsection{Segmentation Performances-ssTEM data:} \label{S:RESULT_TEM}

\begin{figure}
\begin{center}
\subfigure{\includegraphics[width=0.35\columnwidth, height=0.32\columnwidth]{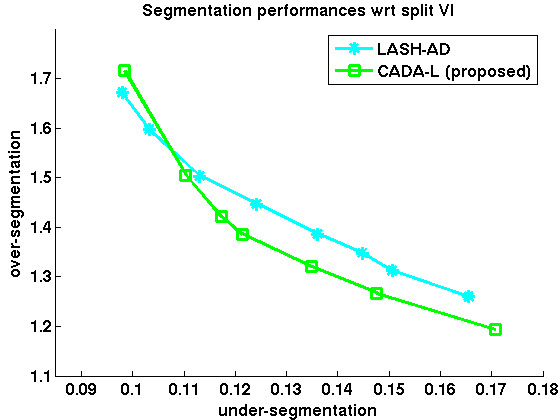}} \qquad
\subfigure{\includegraphics[width=0.35\columnwidth, height=0.32\columnwidth]{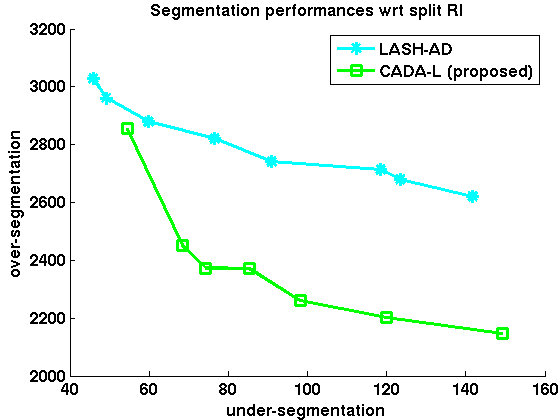}}
\end{center}
\vspace{-0.4cm}\caption[\bf Segmentation errors in terms of split-VI and split-RE  on TEM data.]{\scriptsize Segmentation errors on TEM data. Left column shows split-VI error: $VI_{UE}$ in x-axis, $VI_{OE}$ in y-axis;  right column shows split-RE: $RE_{UE}$ in x-axis, $RE_{OE}$ in y-axis. The curves are averages of errors on 15 $1000 \times 1000$ images. The results of Global method~\cite{andres08} were too poor to plot.}
\label{F:RESULT_TEM}
\end{figure}

This section reports the 2D segmentation results our method and others produced on a different data modality, namely ssTEM images. These images were part of those generated for the work of~\cite{takemura13} and were collected from the authors. Fifteen $500 \times 500$ images were used for training both the pixel and superpixel boundaries. We follow the techniques and 2D versions of features described in Section~\ref{S:OVERSEG} for ssTEM data. The same pixel prediction and watershed regions are provided as input to all competing methods. The segmentation is performed on each image (without connecting them across planes) and Figures~\ref{F:RESULT_TEM}  plots the average of split-VI and split-RE errors over 15 images of size $1000 \times 1000$ of the proposed CADA-L and GALA~\cite{nunez13} methods.  Our method CADA-L seems to produce less over-segmentation in almost all threshold values than that results of GALA~\cite{nunez13}. The result of the Global method~\cite{andres12} were too poor to show on this plot -- lowest over-segmentation error at 4.11 with 0.13 under-segmentation average.

\begin{figure}
\vspace{-1.5cm}
\begin{center}
\includegraphics[width=\columnwidth, height=0.9\columnwidth]{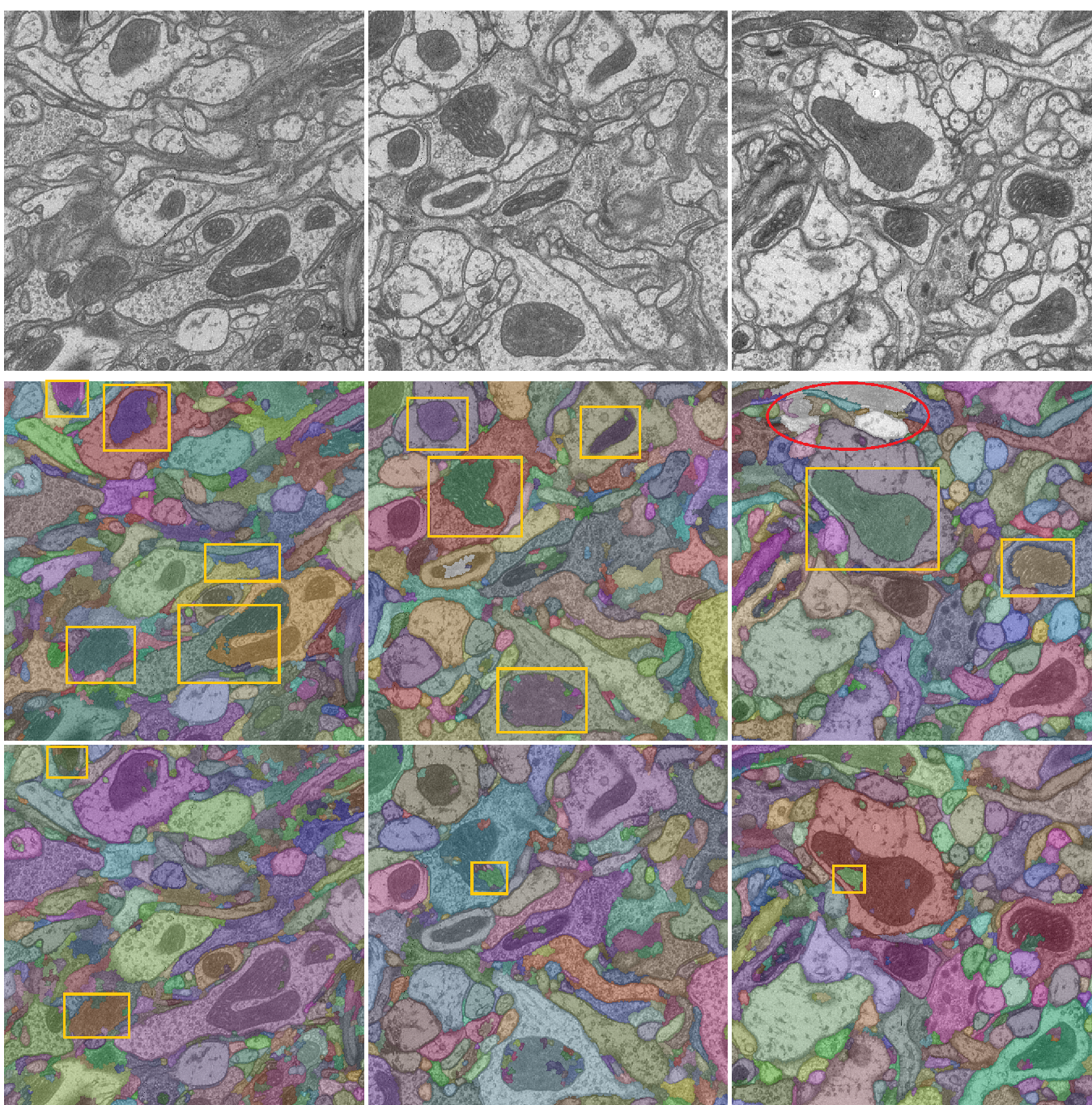}
\vspace{-0.3cm}
\end{center}
\vspace{-0.4cm}\caption[\bf Qualitative comparison on TEM data.]{\scriptsize  Qualitative comparison of GALA and proposed method segmentation on TEM data. The segmentation outputs are overlaid with random colors on the grayscale images. Top row: input, middle GALA and bottom: proposed CADA-L. Significant over-segmentation errors and under-segmentation errors are marked in yellow rectangles and red ellipses respectively.}
\label{F:RESULT_TEM_QUAL}
\end{figure}

In Figure~\ref{F:RESULT_TEM_QUAL}, we show input images and the segmentation results (overlaid on the image) of GALA and our methods at the same under-segmentation error. Examining the qualitative output in Figure~\ref{F:RESULT_TEM_QUAL}, GALA seems to struggle to absorb the mitochondria regions despite multiple learning iterations and even merges two cells in one occasion.   While a more accurate mitochondria predictor could potentially reduce the segmentation errors of the proposed method, context-invariant algorithms such as GALA would be less effective around mitochondria regions. It is worth mentioning here that, compared to GALA, CADA-L used less than half of the training examples ($42.46\%$) collected without training iterations (i.e., significantly more efficient in training).

\section{Discussion}\label{S:DISCUSSION}

We argue that, due to considerable ambiguity in appearances, it is only rational for an EM segmentation algorithm to be context-aware in each of its stages, i.e., in both pixel and superpixel levels (and in alignment for anisotropic data). The results reported in this paper support our claim that a context-aware clustering of sub-classes such as cytoplasm and mitochondria can improve segmentation accuracy significantly given fairly accurate sub-class detection. Our examination of both isotropic and anisotropic data suggests cell structures cannot be meaningfully identified without mitochondria regions and it is non-trivial to combine detection with a segmentation that ignores it (e.g.,~\cite{andres12}) in order to produce the final segmentation . Our analysis also illustrates how a delayed agglomerative procedure benefits from the intermediate boundary probabilities and improves the efficiency of the segmentation process significantly. 



In addition to reducing the over- and under-segmentation errors, one of the variants of our classifier, namely CADA-L, demands substantially fewer training examples (and no training iterations) than others, i.e., its training is significantly faster than others. A context-oblivious strategy gain significantly (compare LASH-D with GALA in Figure~\ref{F:RESULT_Q}) by accumulating training set over multiple iterations.  However, in context-aware approach, one does not benefit much by accumulating the training set (CADA-F in Figure~\ref{F:RESULT_Q}) over a classifier trained from a single iteration (CADA-L). One possible explanation is that previous context oblivious strategies require the extra iterations to mitigate the impact of the noise introduced by mitochondria superpixels. This explanation implies that detecting the sub-classes, and considering them separately as necessary, is perhaps the key to train a boundary classifier accurately and  efficiently. 

We further investigated this conjecture and developed a semi-supervised active learning algorithm (interactive) to train the supevoxel boundary classifier with as few as $< 20\%$ of the total examples~\cite{miccai14interactive}. The requirement of exhaustive labels is a critical bottleneck for automatic EM segmentation, especially for reconstructing larger brain regions, or whole animal brain, where one may anticipate the necessity to train several different classifiers~\cite{helmstaedter13}. The interactive training of both pixels (using Ilastik~\cite{ilastik11} for example) and superpixel boundaries (using~\cite{miccai14interactive}) holds the promise of removing the need for such complete groundtruth and paves the for scaling up the EM reconstruction algorithms. 

We have applied our context aware algorithm to segment 216 FIBSEM volumes of $520^3$ voxels each, with a 10nm isotropic resolution, from the Medulla region of fly retina. As far as we know, this is an attempt to reconstruct one of the largest volumes for such animal. Compared to the result of~\cite{nunez13} on two of the $520^3$ blocks, our segmentation resulted in an estimated $\sim 30\%$ reduction in subsequent manual correction time. In addition, our segmentation were sufficiently accurate for regions that pertains to Post-synaptic densities (PSD), i.e., the synaptic partners of a cell. During the manual annotation of these PSDs, the output of our segmentation method assisted the experts to improve their performances~\cite{plaza14synapse}.


\section{Acknowledgments}
The authors wish to thank Stuart Berg and Bill Katz for their contributions in software development; Pat Rivlin and Shinye Takemura for providing the annotated dataset of~\cite{takemura13} and for the discussion on the neurobilogical properties of our dataset; Matt Saunders and Lei-Ann Chnag for their effort in generating pixel classification ilps and groundtruth annotation; Don Olbris for his assistance with the visualization software Raveler. 


%
%
%

\section{Supporting Information}

%
 
\begin{description}
\item {\bf Code:} \url{https://github.com/janelia-flyem/NeuroProof.git.}

\item {\bf Output volumes:} We have uploaded the grayscale maps and the output labels of several methods at: 
\url{https://www.dropbox.com/sh/1l3eq9c34s7dj68/AAABQP8FZa-AytzZUptHZ2sFa?dl=0.} 

There is also a python code to visualize the results. The format for calling the code is: python ShowSeq.py {\it output-label-path.h5}~~stack~~{\it path-to-input-images}/grayscale-maps/*.png. In addition to pushbuttons, the following keys also work: u -up, d-down, f-toggle the color overlay.

\end{description}

\end{document}